\pdfoutput=1

\documentclass[11pt]{article}
\usepackage[final]{acl}

\usepackage{times}
\usepackage{latexsym}
\usepackage{amsmath}
\usepackage{algorithm}
\usepackage{algpseudocode}
\usepackage{subcaption}
\usepackage{booktabs}
\usepackage{graphicx}
\usepackage{adjustbox}
\usepackage{pdfpages}
\usepackage{placeins}
\usepackage{cleveref}

\usepackage[T1]{fontenc}

\usepackage[utf8]{inputenc}

\usepackage{microtype}

\usepackage{inconsolata}

\usepackage{graphicx}

\title{Text Compression for Efficient Language Generation}

\author{David Gu \\
  ETH Zurich \\
  \texttt{david.gu@inf.ethz.ch} \\
   \\\And
  Peter Belcak \\
  NVIDIA \\
  \texttt{pbelcak@nvidia.com} \\
   \\\And
  Roger Wattenhofer \\
  ETH Zurich \\
  \texttt{wattenhofer@ethz.ch}}

\begin{document}
\maketitle

\begin{abstract}
We challenge the prevailing assumption that LLMs must rely fully on sub-word tokens for high-quality text generation. To this end, we propose the ``Generative Pretrained Thoughtformer'' (GPTHF), a hierarchical transformer language model capable of text generation by compressing text into sentence embeddings and employing a sentence attention mechanism. GPTHF retains GPT’s architecture, modifying only token interactions via dynamic sparse attention masks. 

Our experiments show that GPTHF achieves an up to an order of magnitude improvement in FLOPs efficiency and a threefold increase in runtime speed compared to equally-sized GPT models in the low-size regime. This is achieved through a unique generation method that caches and reuses sentence embeddings, allowing significant portions of the input to bypass large parts of the network.
\end{abstract}

\section{Introduction}
The development of LLMs has garnered substantial interest due to their impressive capabilities in NLP tasks. The dominant paradigm for improving LLMs has been \textit{scaling}, with models scaling from hundreds of millions (e.g. BERT, \citet{devlin2018bert}) to over a trillion parameters (e.g. Switch Transformer, \citet{fedus2022switch}) in a span of four years. While these massive scales unlock remarkable performance across NLP tasks \citep{naveed2023comprehensive}, they come with substantial costs in hardware, energy, and time \citep{strubell2019energy, patterson2021carbon}, requiring the exploration for more efficient methods.

Efforts to improve efficiency include pruning \citep{augasta2013pruning}, quantization \citep{hubara2018quantized}, and knowledge distillation \citep{gou2021knowledge}. Mixture of experts models \citep{shazeer2017outrageously, fedus2022switch} further reduced inference costs while preserving capacity. However, one area remains under-explored: the reliance of LLMs on sub-word tokens, each requiring embeddings several kilobytes in size. This raises the question of whether more condensed text representations could offer similar performance with greater efficiency. Models like the Funnel-Transformer \citep{dai2020funnel} hint at potential gains through compressing and subsequently decompressing hidden states.

Going one step further, we introduce GPTHF, a hierarchical transformer that compresses entire sentences into fixed-size embeddings. We explore whether such representations still carry sufficient semantic payload to maintain generation quality, thereby asking if sub-word tokens could possibly be eliminated for greater computational efficiency. Experimental results show that GPTHF achieves strong perplexity scores, follows scaling laws in the low-parameter regime, and operates at a significantly reduced FLOPs cost and inference time.

\paragraph{Contributions.}
1. We propose GPTHF, a transformer language model that generates text by compressing sentences into one fixed-size embedding and employing sentence-level attention, with minimal modifications to GPT. 2. We introduce a generation method that caches and reuses sentence embeddings, yielding linear efficiency improvements with context size, achieving up to 10x FLOP reductions and 3x runtime speedup. 

\section{Related Work}
A new line of research explored the idea of a ``hierarchical transformer,'' a transformer operating on variable-size embeddings within different layers of the network. Early examples include the models of \citet{yang2016hierarchical} and \citet{montero2021sentence}. The Funnel Transformer \citep{dai2020funnel} compressed token sequences via incremental pooling, with inter-layer skip connections allowing later layers to access pre-compressed information. When re-investing the saved FLOPs, the Funnel Transformer outperformed previous state-of-the-art models with comparable computational resources. \citet{nawrot2021hierarchical} expanded this idea to generative transformers with their ``Hourglass'' model, demonstrating improved perplexity on a Wikipedia dataset. Other examples include Sentence-BERT \citep{reimers2019sentence} and Sentence-GPT \citep{muennighoff2022sgpt}, focus on generating sentence embeddings for downstream tasks.

Our work differs from all of the above in several ways. Instead of compressing a fixed-size group of tokens, we compress a sentence -- a unit of higher semantic value in language -- into one embedding. We focus on leveraging these embeddings to improve computational efficiency, not on the embeddings themselves.

\section{Methodology}
\subsection{Architecture}
The GPTHF model consists of two main components: a word-level transformer encoder (\texttt{wlt\_encoder}) and a sentence-level transformer body (\texttt{slt\_body}). The encoder compresses each sentence into a single embedding while preserving essential information. The \texttt{slt\_body} contextualizes these sentence embeddings and generates the next-token prediction.

During the forward pass (see \Cref{fig:generative_inf}), the input tokens $x_1,\cdots,x_n$ are first processed by the \texttt{wlt\_encoder}, producing contextualized sub-word embeddings. The \texttt{wlt\_encoder} uses block attention masks, which will be explained below. Fetching the last token of each sentence $s_i$ yields an embedding $e_i, i \in [m]$:
\begin{equation*}
    e_i=\text{Pooling}(\texttt{wlt\_encoder}(x_1,...,x_n)),
\end{equation*} where $m$ is the number of sentences. These embeddings are then processed by the \texttt{slt\_body}:
\begin{equation*}
    \hat{e_i}=\texttt{slt\_body}(e_1,...,e_n)), i \in [m].
\end{equation*} Finally, $\hat{e_m}$ is fed into the language modeling head to predict the next token.

\begin{figure}[ht]
    \centering
    \begin{adjustbox}{center,max width=\textwidth}
        \includegraphics[trim=20 370 610 50, clip, width=\linewidth]{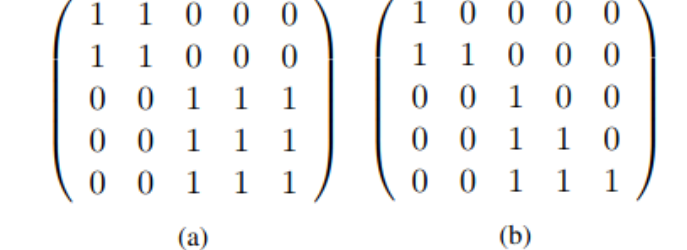}
    \end{adjustbox}
    \caption{Visualization of block attention masks for a text with sentence index vector $[0,0,1,1,1]$. (a) A block matrix allowing attention within sentences. (b) Block lower triangular matrix allowing attention to previous tokens within sentences during training.}
\label{fig:block-masks}
\end{figure}

\begin{figure}[ht]
    \centering
    \begin{adjustbox}{center,max width=\textwidth}
        \includegraphics[trim=20 150 220 60, clip, width=\linewidth]{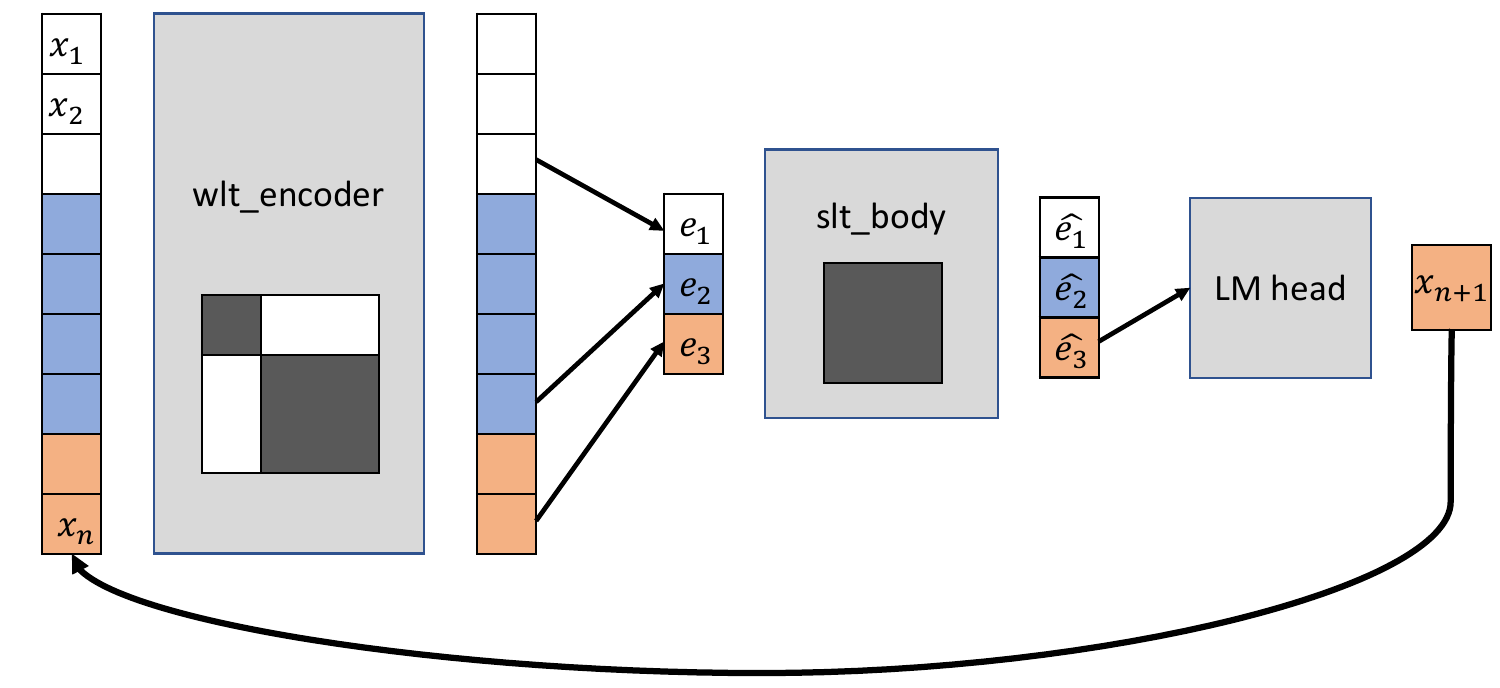}
    \end{adjustbox}
    \caption{Overview of the Generative THF (GPTHF) Architecture during inference. The boxes in the models indicate the type of attention masks used. The attention masks are explained in \Cref{fig:block-masks}.}
    \label{fig:generative_inf}
\end{figure}

\paragraph{Block attention masks.}
\label{sec:block_masks}
To ensure sentence embeddings capture only intra-sentence information, we use a localized attention mechanism that restricts token attention to within the same sentence. This is enforced via a dynamically computed (for each input) block attention mask, defined by a \textit{sentence index vector} at tokenization time. Each block corresponds to a sentence, preventing cross-sentence interactions (see \Cref{fig:block-masks}).

\begin{table}[ht]
    \centering
    \resizebox{0.49 \textwidth}{!}{%
        \begin{tabular}{lccccccc}
            \toprule
            Name & Params & $d$ & $n_{heads}$ & $l_{enc}$ & $l_{body}$ & $lr$ \\
            \midrule
            GPTHF-8-4 & 151M & 768 & 12 & 8 & 4 & 6e-4 \\
            GPTHF-16-8 & 454M & 1024 & 16 & 16 & 8 & 4e-4 \\
            \bottomrule 
        \end{tabular}
    }
    \caption{Model sizes and hyperparameters for GPTHF models.}
    \label{tab:generative_hyperparams}
\end{table}

\paragraph{Model sizes and Details.}
A summary of the model sizes and other hyperparameters are provided in \Cref{tab:generative_hyperparams}. Through empirical experimentation, a relatively large encoder is found beneficial. We decide on the following modifications over the vanilla transformer \citep{vaswani2017attention}, mostly inspired by Llama-1 \citep{touvron2023llama} and \citet{geiping2023cramming}, who proposed architectural changes when training language models in low-compute settings. 

First, we replace an absolute positional embedding layer with rotary positional embeddings (RoPE, \citet{su2024roformer}) at each attention layer of the network. We use SwiGLU activation \citep{shazeer2020glu} with a dimension of 2/3 4d. Moreover we use pre-normalization layers with RMSNorm \citep{zhang2019root}. Finally, we disable all QKV biases in the transformer attention layers and linear layers.

\subsection{Pre-training} 
\label{sec:generative_train}
\begin{figure}
    \centering
    \begin{adjustbox}{center,max width=\textwidth}
        \includegraphics[trim=20 160 280 60, clip, width=\linewidth]{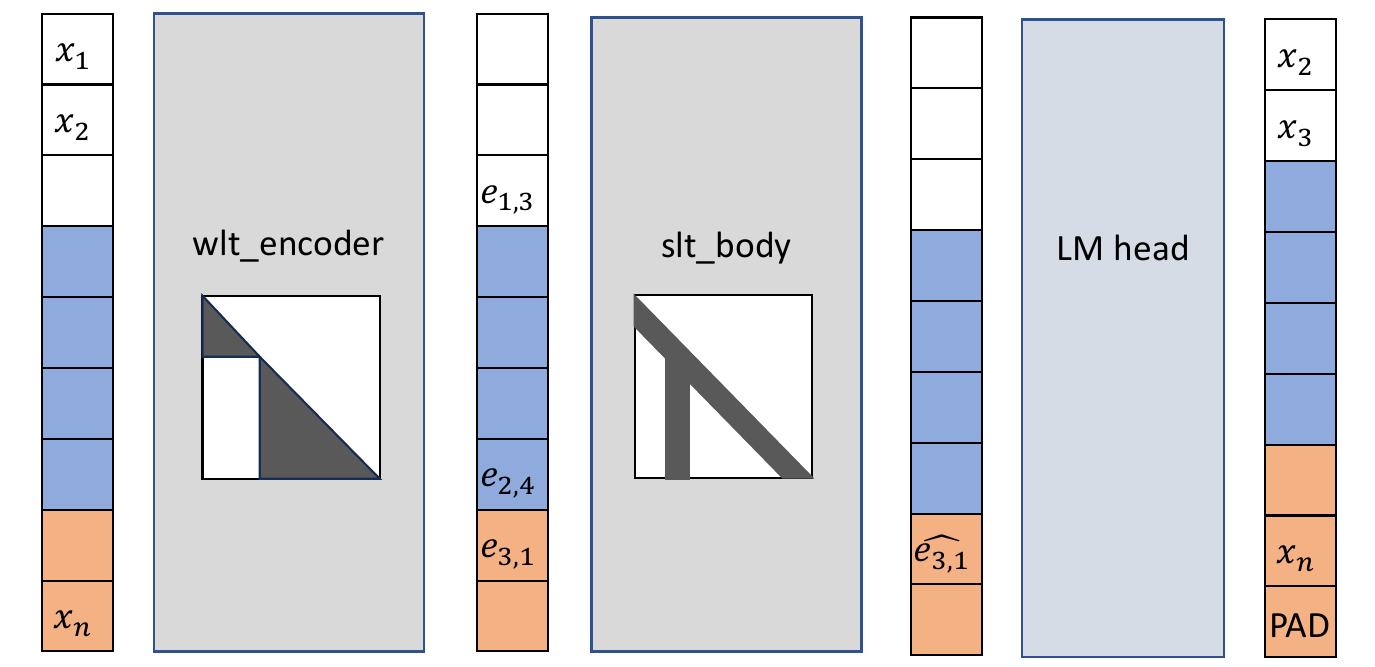}
    \end{adjustbox}
    \caption{Overview of the pre-training procedure. The boxes in the models indicate the type of attention masks used. The attention masks are explained in \Cref{fig:generative_train_masks}.}
    \label{fig:generative_train}
\end{figure}

\begin{figure}
    \centering
    \begin{adjustbox}{center,max width=\textwidth}
        \includegraphics[trim= 20 350 610 40, clip, width=\linewidth]{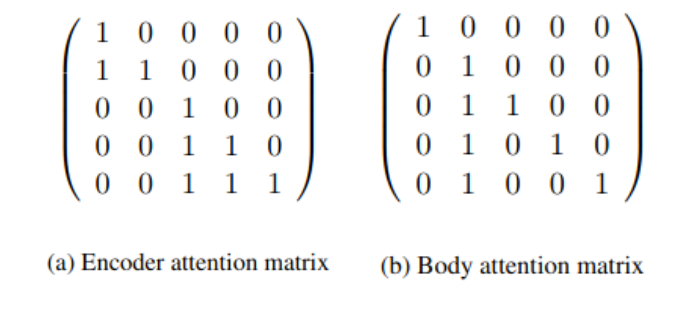}
    \end{adjustbox}
    \caption{Attention masks during pre-training for an input with the sentence index vector [0,0,1,1,1]: The left matrix is the "block triangular mask" as in \Cref{sec:block_masks}. After going through the encoder, every token represents the compressed prefix of its sequence up to itself, and is only allowed to attend to itself and compressions of previous sequences (right).}
\label{fig:generative_train_masks}
\end{figure}

We use the next token prediction objective common in auto-regressive models. To prepare GPTHF for token prediction while enabling efficient parallel training, we again employ specialized attention masks (\Cref{fig:generative_train_masks}). The target is the next token in the sequence (\Cref{fig:generative_train}).

Interestingly, training GPT and GPTHF differs only in replacing full triangular attention matrices with dynamically computed sparse ones, with no architectural changes.

\paragraph{Data.}
Our training corpus incorporates OpenWebText, Wikipedia and ArXiv. OpenWebText forms the backbone due to its large size and diverse internet content. Wikipedia is known for its vast coverage of general knowledge. Finally, ArXiv augments our corpus with scientific and technical texts. We use the standard GPT-2 tokenizer, inheriting its handling of vocabulary size and unknown words, while introducing an ``end-of-sentence'' token. This token is crucial in the design of a fast generation method, a cornerstone of this work.

\paragraph{Details.}
We use the Adam optimizer with weight decay of 0.01, $\beta_1 = 0.9$, $\beta_2 = 0.98$ and $\epsilon = 10^{-8}.$ We maintain gradient clipping with a value of 0.5. As our learning rate scheduler we use linear decay with 10000 warmup steps. The peak learning rates are provided in \Cref{tab:generative_hyperparams}. We keep the batch size scheduler from \cite{geiping2023cramming}, starting batch size at 64 and linearly ramping up to 4096, reaching this peak at 60\% of the training duration. Lastly, we eliminate dropout during training. Our models undergo only a single pass or less over the pre-training corpus, which mitigates the risk of overfitting.

\subsection{Fast generation}
\label{sec:fast_gen}
\begin{figure}[ht]
    \centering
    \begin{adjustbox}{center,max width=\textwidth}
        \includegraphics[trim=120 30 70 30, clip, width=\linewidth]{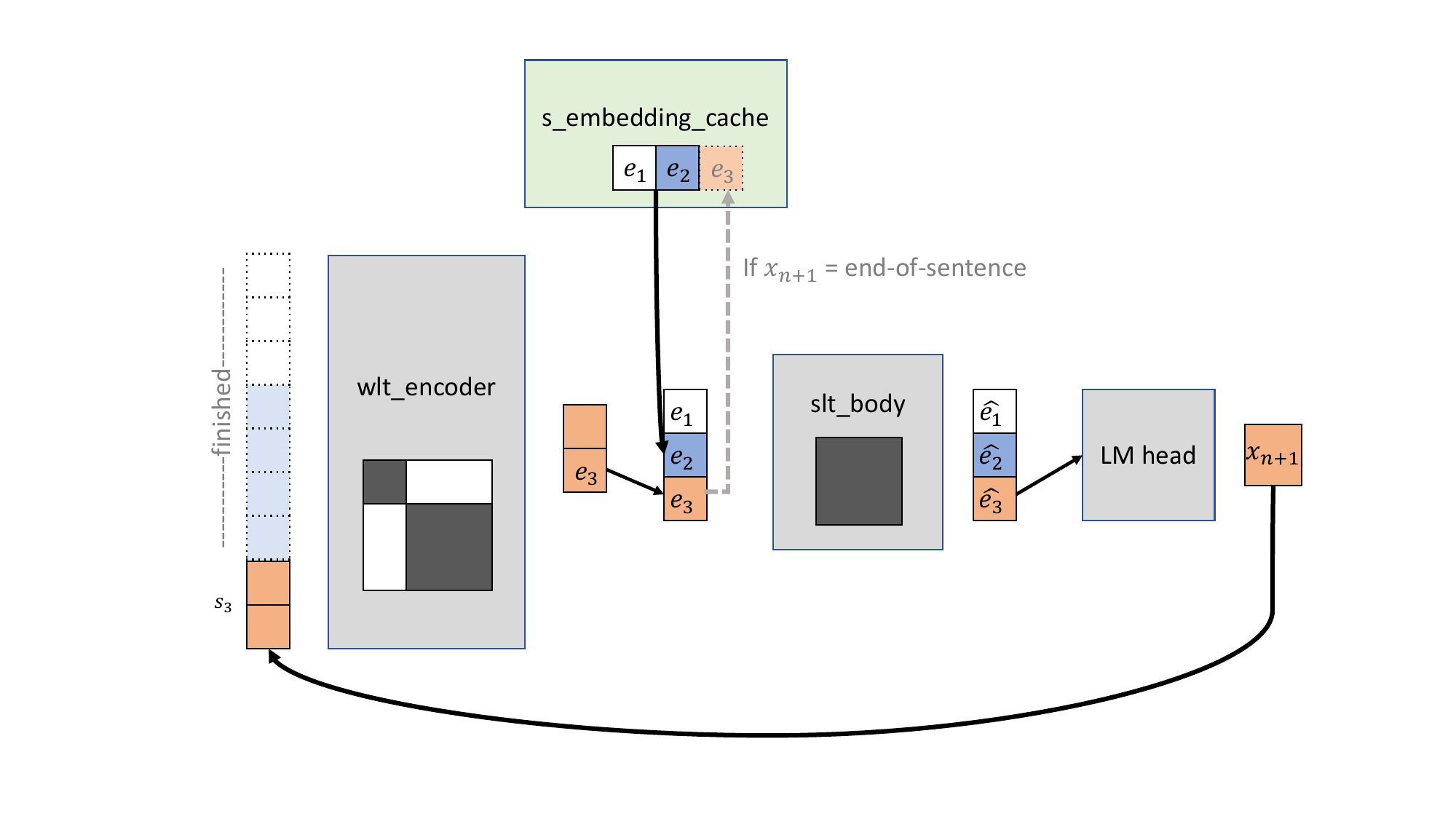}
    \end{adjustbox}
    \caption{Illustration of the Fast Generation Algorithm. Having finished $s_1$ and $s_2$ in the context, any subsequent token mathematically cannot influence $e_1, e_2$. The Fast Generation Algorithm caches them and feeds them directly to the \texttt{slt\_body}, together with $e_3$.}
    \label{fig:fast_gen}
\end{figure}

The insight that enables a faster generation algorithm to be mathematically equivalent to regular token generation is the design of our block-wise attention matrix. During the generation loop, when generating a token in sentence $j$, only tokens in sentence $j$ are affected -- tokens in previous sentences remain unchanged. Since the feed-forward layers operate element-wise, there is no operation within the transformer layer that alters the compressed embeddings $e_1, e_2,\cdots, e_{j-1}$. The core idea is to cache these embeddings, allowing the encoder to process only the current sentence $j$ to compute $e_j$.  The body then processes the concatenation of the cached embeddings $e_1, e_2,\cdots, e_{j-1}$ and the updated $e_j$. For an illustration, see \Cref{fig:fast_gen}.

\section{Experiments}
\subsection{Setup}
We evaluate GPTHF against GPT-style baselines of comparable size, using validation perplexity and efficiency metrics (FLOPs and runtime). Due to computational constraints, the training data is limited to 10 billion tokens, divided into 320'000 micro-batch steps of size 64 with a context size of 512 tokens. All models are pre-trained on the same datasets.

\paragraph{Baselines.}
We trained a 12-layer baseline named ``Baseline-12'' and a 24-layer ``Baseline-24'' with the same architecture and size as their GPTHF counterparts. The only difference was that they were trained using full triangular masks for both encoder and body, as opposed to the masks in \Cref{fig:generative_train_masks}. As remarked in \Cref{sec:generative_train}, the baselines can be regarded as equivalent to conventional GPTs.

\subsection{Perplexity}
\begin{figure}
    \centering
    \includegraphics[trim=0 0 0 0, clip, width=\linewidth]{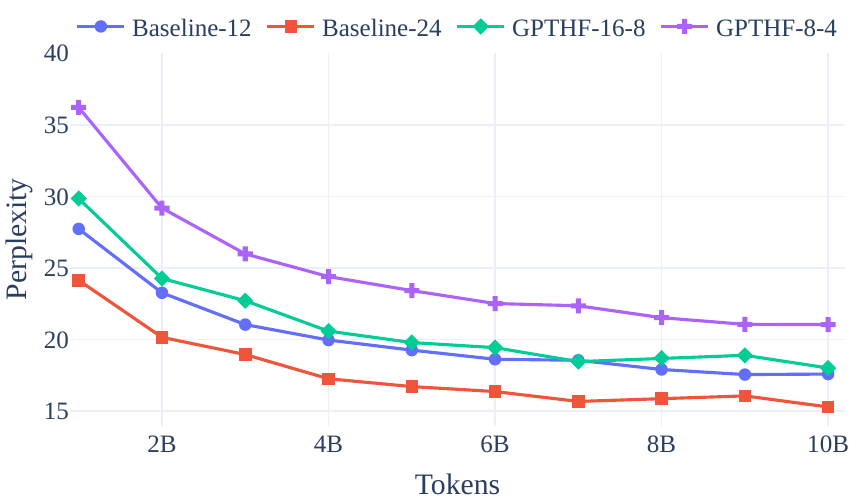}
    \caption{Validation perplexity of pre-trained models and baselines. Lower values indicate better performance.}
    \label{fig:val_perplexities}
\end{figure}

Validation perplexities after training are presented in \Cref{fig:val_perplexities}. They were calculated on a hold-out validation dataset comprising 16 million tokens.

\paragraph{Scaling Laws Hold in the Low-Compute Setting.}
GPTHF models have higher perplexity than baselines but follow scaling laws in the low-parameter regime. Both show a  $\sim$5-point perplexity drop when scaling from 12 to 24 layers after 10B tokens. GPTHF-16-8 and the 12-layer baseline perform on par, setting a basis for further comparisons: If GPTHF-16-8 achieves higher generation efficiency and/or speed than a 12-layer GPT, training a larger model capable of compression might be worthwhile.

\subsection{FLOPs}
The speedup from our fast generation algorithm (\Cref{sec:fast_gen}) depends on token distribution across sentences as opposed to only the shape of the input. Intuitively, more sentences help by caching completed ones to skip the encoder. Since theoretical FLOPs analysis is impractical, we measure empirically using OpenWebText samples with varying prompt lengths ($n$) and token counts ($k$), leveraging the tool from \citet{flopsprofiler}. All numbers in \Cref{tab:flops} exclude KV-caching \citep{pope2023efficiently}, as adapting our approach to it requires significant additional effort.

\begin{table*}[ht]
    \centering
    \resizebox{\textwidth}{!}{%
    \begin{tabular}{lccccc|ccccc}
        \toprule
        & \multicolumn{5}{c|}{\textbf{Batch size 1}} & \multicolumn{5}{c}{\textbf{Batch size 32}} \\
        \midrule
        $n,k=$ & 100,100 & 100,250 & 250,100 & 250,250 & 500,20 & 100,100 & 100,250 & 250,100 & 250,250 & 500,20 \\
        \midrule
        Baseline-12 & 2.38T & 9.1T & 4.88T & 15.7T & 1.56T & 2.46T & 9.62T & 4.96T & 16.0T & 1.7T \\
        GPTHF-8-4 & 0.95T & 4.16T & 0.80T & 4.31T & 0.17T & 1.90T & 7.72T & 2.53T & 9.32T & 0.58T \\
        Efficiency & 2.51x & 2.19x & 6.10x & 3.64x & \textbf{9.18x} & 1.29x & 1.25x & 1.96x & 1.72x & \textbf{2.93x} \\
        \midrule
        Baseline-24 & 8.30T & 31.4T & 17.0T & 53.9T & 5.45T & 8.52T & 32.7T & 17.2T & 54.9T & 5.95T \\
        GPTHF-16-8 & 2.99T & 17.4T & 2.97T & 17.5T & 0.56T & 6.11T & 25.6T & 8.39T & 31.3T & 2.04T \\
        Efficiency & 2.78x & 1.81x & 5.72x & 3.08x & \textbf{9.73x} & 1.39x & 1.28x & 2.05x & 1.75x & \textbf{2.92x} \\
        \bottomrule
    \end{tabular}
    }
    \caption{Empirical FLOP count per sample for varying prompt lengths $n$ and generated token counts $k$. Lower values indicate better efficiency. Bold values highlight highest speedup for each batch size. The mean over 50 batches is reported. Efficiency is calculated as the inverse of the FLOP reduction of the GPTHF model compared to its respective baseline.}
    \label{tab:flops}
\end{table*}

\paragraph{Efficiency Gain Increases With Prompt Length.}
The results show that efficiency improves with larger $n$, but surprisingly decreases with higher $k$. A closer examination reveals that our models generate few relevant tokens, often repeating them without generating end-of-sentence tokens. This occurs in both GPTHF models and baselines, indicating that it likely stems from insufficient scale or training rather than compression. Since the fast algorithm relies on completed sentences, generation quality directly affects efficiency. This explains a) the small gains 100-prompt/250-generation tokens, and b) strong efficiency gains (up to 10x) for 500-prompt/20-generation tokens. We hypothesize that a model capable of correctly terminating sentences achieves greater efficiency gains than reported in \Cref{tab:flops}, increasing with both $n$ and $k$.

\paragraph{Sentences vs Efficiency.}

\begin{figure}[htbp]
    \centering
    \begin{subfigure}[b]{0.49\linewidth}
        \includegraphics[trim=0 20 0 30, clip, width=\linewidth]{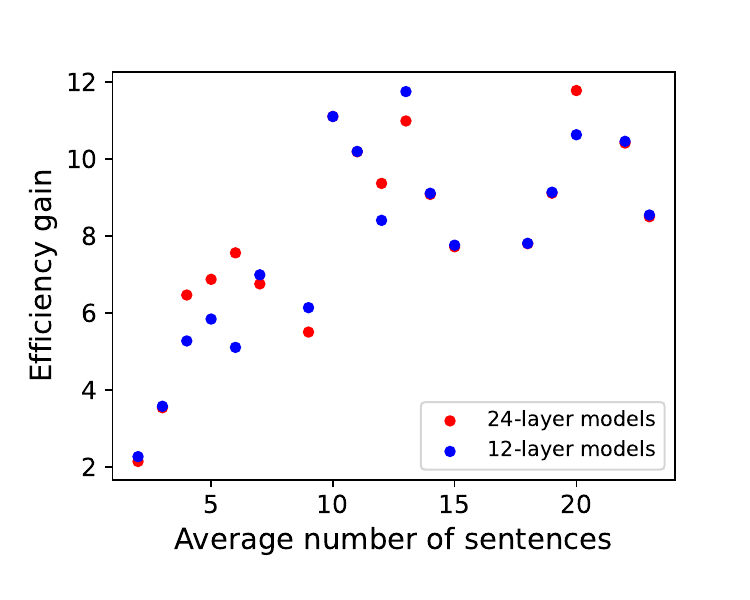}
        \caption{Batch size = 1}
        \label{fig:efficiency_1}
    \end{subfigure}
    \begin{subfigure}[b]{0.49\linewidth}
        \includegraphics[trim=0 20 0 30, clip, width=\linewidth]{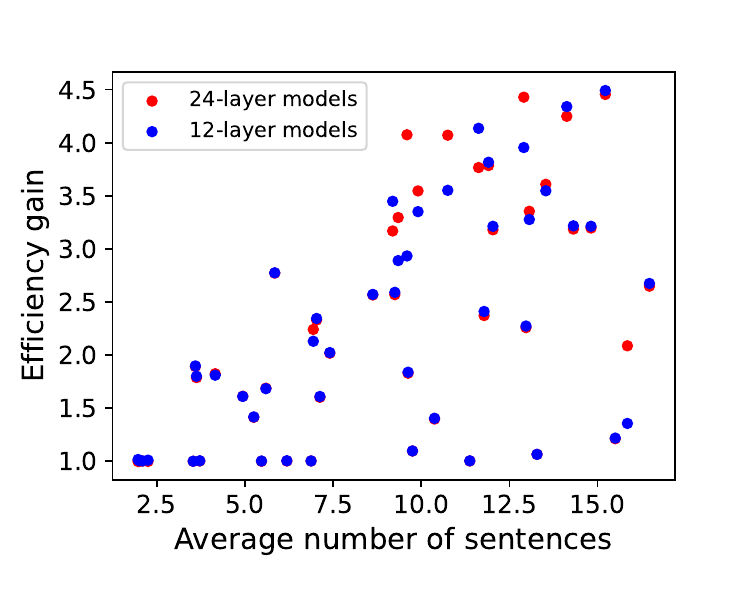}
        \caption{Batch size = 32}
        \label{fig:efficiency_32}
    \end{subfigure}
    \caption{Scatter plots showing the average number of sentences (x-axis) versus the efficiency gain (y-axis) of GPTHF over GPT when generating 20 tokens.}
    \label{fig:scatter_speedup}
\end{figure}

\Cref{fig:scatter_speedup} shows scatter plots of the average sentence count (x-axis) versus efficiency gain (y-axis). We see that the efficiency gain increases \textit{linearly} with the average number of sentences. For batched data, the efficiency gain is lower likely due to larger variety (which can be observed from the increased variance) in tokens, leading to more padding tokens being processed, which slows the fast generation algorithm.

\subsection{Inference Time}
While we save many FLOPs, not all translate to faster runtime due to GPU inefficiencies from non-trivial and conditional executions. We measure actual inference times to account for this, using an identical setup.

\begin{table*}[ht]
    \centering
    \resizebox{\textwidth}{!}{%
    \begin{tabular}{lccccc|ccccc}
        \toprule
        & \multicolumn{5}{c|}{\textbf{Batch size 1}} & \multicolumn{5}{c}{\textbf{Batch size 32}} \\
        \midrule
        $n,k=$ & 100,100 & 100,250 & 250,100 & 250,250 & 500,20 & 100,100 & 100,250 & 250,100 & 250,250 & 500,20 \\
        \midrule
        Baseline-12 & 1.73s & 4.44s & 1.82s & 4.77s & 0.44s & 0.17s & 0.57s & 0.28s & 0.88s & 0.093s \\
        GPTHF-8-4 & 1.77s & 4.46s & 1.77s & 4.48s & 0.41s & 1.90T & 0.50s & 0.18s & 0.56s & 0.041s \\
        Efficiency & 0.98x & 1.00x & 1.03x & 1.06x & \textbf{1.07x} & 1.13x & 1.14x & 1.56x & 1.57x & \textbf{2.27x} \\
        \midrule
        Baseline-24 & 3.40s & 8.88s & 3.73s & 9.85s & 0.84s & 0.40s & 1.42s & 0.73s & 2.34s & 0.26s \\
        GPTHF-16-8 & 3.32s & 8.43s & 3.32s & 8.44s & 0.67s & 0.35s & 1.24s & 0.37s & 1.29s & 0.087s \\
        Efficiency & 1.02x & 1.05x & 1.12x & 1.17x & \textbf{1.25x} & 1.14x & 1.15x & 1.97x & 1.81x & \textbf{2.99x} \\
        \bottomrule
    \end{tabular}
    }
    \caption{Empirical generation time in seconds per sample for different prompt lengths $n$ and number of tokens generated $k$. Lower values are better. Bold values indicate highest speedup for each batch size. The mean over 50 batches executed on a single NVIDIA RTX A6000 is reported. Speedup is calculated as the inverse time reduction of our model in comparison to the baseline.}
    \label{tab:speed}
\end{table*}

\paragraph{Speedup Increases With Context.}
Similar to the FLOP experiment, increasing up to 25\% for unbatched data as $k$ grows. Batched data shows gains with larger $n$ but not $k$, which we attribute to the same sentence-termination limitations.

\paragraph{Latency vs. Throughput.}
We attribute the significant speedup differences between unbatched and batched data to latency vs. throughput. For unbatched data with small contexts, the GPU remains idle. This limits the runtime by latency, which primarily depends on model size. Batched data utilizes GPUs better, converting efficiency gains in FLOPs into higher throughput. Moreover, speedup increases with model size, resulting in up to triple the speedup when comparing GPTHF with equal-sized baselines and slightly faster when comparing GPTHF 16-8 with the 12-layer baseline.

\paragraph{Sentences vs. Speedup.}

\begin{figure}[htbp]
    \centering
    \begin{subfigure}[b]{0.49\linewidth}
        \includegraphics[trim=0 0 0 30, clip, width=\linewidth]{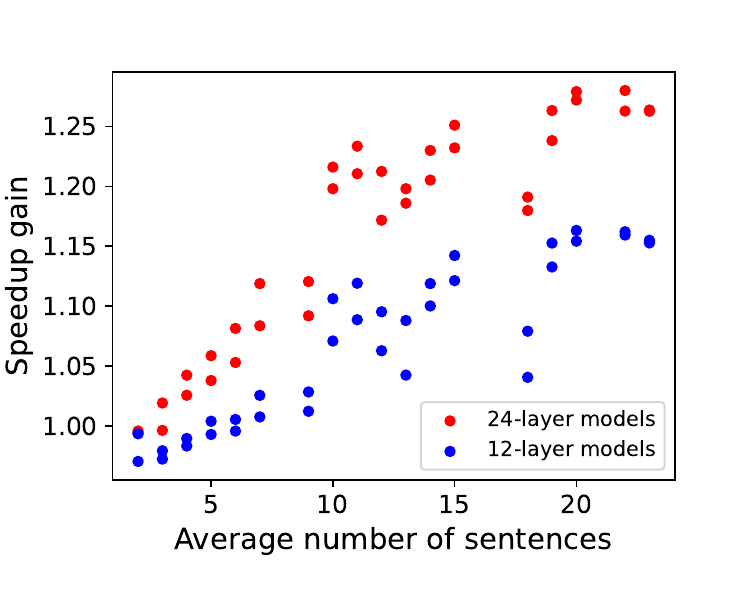}
           \caption{Batch size = 1}
        \label{fig:speedup_1}
    \end{subfigure}
    \begin{subfigure}[b]{0.49\linewidth}
        \includegraphics[trim=0 0 0 30, clip, width=\linewidth]{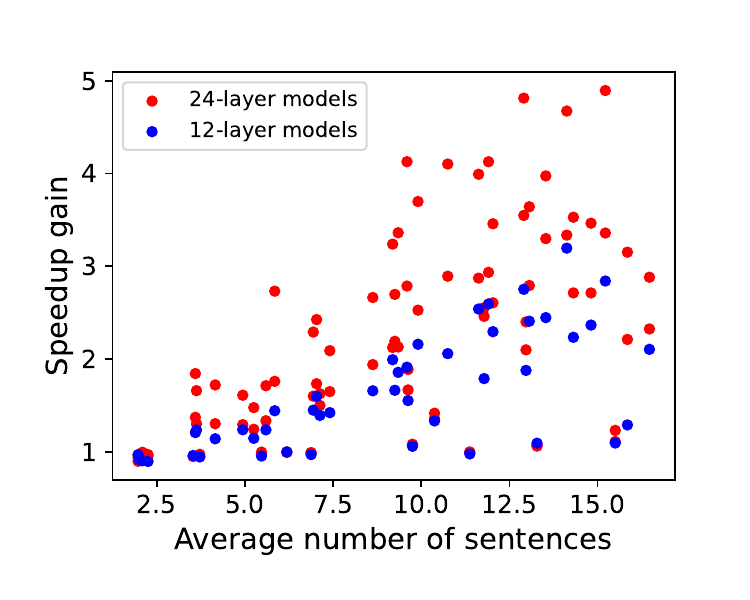}
           \caption{Batch size = 32}
        \label{fig:speedup_32}
    \end{subfigure}
    \caption{Scatter plots showing the average number of sentences (x-axis) versus the speedup gain (y-axis) of GPTHF over GPT when generating 20 tokens.}
    \label{fig:scatter_time}
\end{figure}

\Cref{fig:scatter_time} plots average sentence count (x-axis) against runtime speedup (y-axis). The figure highlights a \textit{linear} relationship between the number of sentences and the speedup, with a larger constant for a larger model size.

\subsection{Discussion}
Our experiments show that compression results in a notable performance drop. Switching from a baseline/GPT to a GPTHF increases perplexity by ~5 points after 10B tokens of training, similar to reducing a 24-layer GPT to 12 layers. 

However, GPTHF models exhibit promising scaling behavior and significant efficiency improvements. Our method achieves speedups of up to 10x in FLOPs and 3x in runtime, scaling linearly with context size. For both our method and the baseline, KV-caching was excluded. Future work might want to explore KV cache integration to evaluate the effectiveness of our approach over state-of-the-art implementations.

Evaluating the overall tradeoff, we compare the GPTHF-16-8 and the 12-layer baseline, which perform on par (\Cref{fig:val_perplexities}). When processing 500 tokens of context, GPTHF-16-8 uses $\sim1/3$ of the FLOPs for unbatched data and is slightly faster (7\%) for batched data. Larger prompt lengths and batch sizes are expected to amplify these gains, making the tradeoff worthwhile at low compute scales.

These results suggest that sentence embeddings \textbf{could replace sub-word tokens in low-compute settings} while maintaining reasonable perplexity, but whether they remain competitive at larger scales is still open. 



\newpage

\section{Limitations}
The central question remains in whether transformers can generate high-quality text using only compressed sentence embeddings with sufficient size and training. While smaller GPTHF models follow scaling laws similar to GPTs, their inability to reliably finish sentences highlights challenges tied to either scale or the compression method itself. Further training on larger models is necessary to determine if this limitation is inherent to compression or surmountable via scaling.

Future work should evaluate these models on downstream tasks to assess practical utility beyond perplexity. Additionally, integrating GPTHF with existing optimizations like KV-caching could yield better speedups, though diminishing returns are a potential challenge. Comprehensive ablation studies focusing on key parameters like hidden size could offer deeper insights into performance. Alternative approaches, such as directly compressing sequences into single embeddings, warrant exploration to enhance or complement current methods.

\end{document}